\documentclass[runningheads]{llncs}
\usepackage[T1]{fontenc}
\usepackage{multicol}
\usepackage{multirow}
\usepackage{graphicx,verbatim}
\usepackage{amsmath,amssymb,amsfonts}
\usepackage{graphicx}
\usepackage{textcomp}
\usepackage{booktabs}
\usepackage{makecell}   
\usepackage{pifont} 
\usepackage{color} 

\begin{document}
\title{RadHiera: Semantic Hierarchical Reinforcement Learning for Medical Report Generation}
%

\author{Bodong Du\inst{1} \and
Honglong Yang\inst{1} \and
Xiaomeng Li\inst{1}\thanks{Corresponding author: eexmi@ust.hk}}

\authorrunning{B. Du et al.}

\institute{The Hong Kong University of Science and Technology (HKUST), Hong Kong SAR, China}

  
\maketitle              
\begin{abstract}
Vision-language models have shown promising results in radiology report generation. However, most existing methods generate reports as flat text and do not explicitly model the semantic dependency between the \textit{Findings} and \textit{Impression} sections, which can lead to inconsistencies between clinical observations and diagnostic conclusions. In this paper, we propose \textbf{RadHiera}, a semantic hierarchical reinforcement learning framework for radiology report generation. \textbf{RadHiera} follows the semantic organization of radiology reports by first optimizing overall report quality, then improving the diagnostic accuracy of the \textit{Impression} section, and finally enforcing consistency between \textit{Findings} and \textit{Impression} so that diagnostic conclusions are supported by clinical evidence. Specifically, we begin with a base reward that combines linguistic quality and medical factuality to provide supervision on the whole report. On this basis, we introduce a severity-aware reward for the \textit{Impression} section that places greater emphasis on errors involving clinically critical conditions, thereby reducing both missed diagnoses and overstatement. We further enforce cross-section consistency using Expert Model-derived label sets, with subset constraints and hallucination penalties to ensure that impressions remain faithful to the findings. Experiments on three public chest X-ray benchmarks show that \textbf{RadHiera} consistently improves diagnostic accuracy and inter-section consistency over state-of-the-art methods, while also demonstrating good adaptability to report generation in ultrasound report generation.

\keywords{Medical Report Generation \and Reinforcement Learning \and Semantic Hierarchy.}

\end{abstract}

\section{Introduction}

Radiology reports adhere to a hierarchical structure where Findings describe observations and Impression synthesizes conclusions, ensuring clinical reliability \cite{schwartz2011improving,ganeshan2018structured}. However, existing MRG systems underexplore this dependency. Early frameworks \cite{jing2017automatic,li2018hybrid,smit2020chexbert,chen2022cross,hou2023recap,wang2022automated,2023_Yan} neglect the Impression section, due to its higher semantic complexity and optimization difficulty. In recent few years, MLLMs \cite{wang2023r2gengptradiologyreportgeneration,medgemma2024,zhou2025medversageneralistfoundationmodel} generate whole reports including Findings section and Impression section but fail to explicitly model the inter-section hierarchy. As a result, generated reports may exhibit diagnostic inaccuracies or inconsistencies, such as missing critical pathologies or introducing unsupported conclusions (see the "MedVersa" results in Figure~\ref{fig:qualitative}).

To address this structural mismatch and improve clinical reliability, we propose \textbf{RadHiera}, a semantic hierarchical reinforcement learning framework that directly optimizes policy behavior toward clinically structured reporting. Central to this approach is a mutli-level reward mechanism that explicitly models the semantic dependency between Findings and Impression sections.
 First, we apply a base reward that integrates linguistic fluency and medical factual correctness, ensuring basic coherent and clinically valid report generation. Building on this, we introduce a \textbf{severity-aware Impression reward} for the Impression section. It compares the generated Impression with the reference and assigns higher weights to clinically critical conditions (e.g., pneumothorax, pulmonary edema). This reward penalizes omissions of high-risk findings and discourages overstatements, thus prioritizing critical diagnoses and improving clinical fidelity. Finally, we introduce an \textbf{cross-sectional consistency reward} between the \textit{Findings} and \textit{Impression} sections, using CheXbert-derived label sets. The reward enforces consistency by applying subset constraints and hallucination penalties, reducing diagnostic contradictions and promoting structured alignment across sections.

Extensive experiments  across three public chest X-ray benchmarks \cite{demner2016preparing,johnson2019mimic,zhang2025rexgradient} demonstrate that RadHiera consistently enhances diagnostic precision and inter-section coherence relative to state-of-the-art baselines. Furthermore, the framework exhibits remarkable cross-modal generalizability, extending its performance gains to the domain of ultrasound report generation in an in-house CarotidUS-MRG dataset.

\noindent\textbf{To summarize, this work makes the following contributions:}
\begin{itemize}
   \item We propose \textbf{RadHiera}, a semantic hierarchical reinforcement learning framework that incorporates clinical constraints into the policy optimization process, explicitly modeling the hierarchical structure of MRG.
    
    \item We design a \textbf{multi-level reward mechanism} composed of a base reward, a severity-aware Impression reward for diagnostic prioritization, and a cross-sectional consistency reward that enforces alignment between report sections.
    
    \item We conduct comprehensive experiments on diverse MRG datasets, demonstrating that our method achieves state-of-the-art performance on MRG
    both quantitatively and qualitatively. 
\end{itemize}

\begin{figure*}[t] \centering \includegraphics[width=\textwidth]{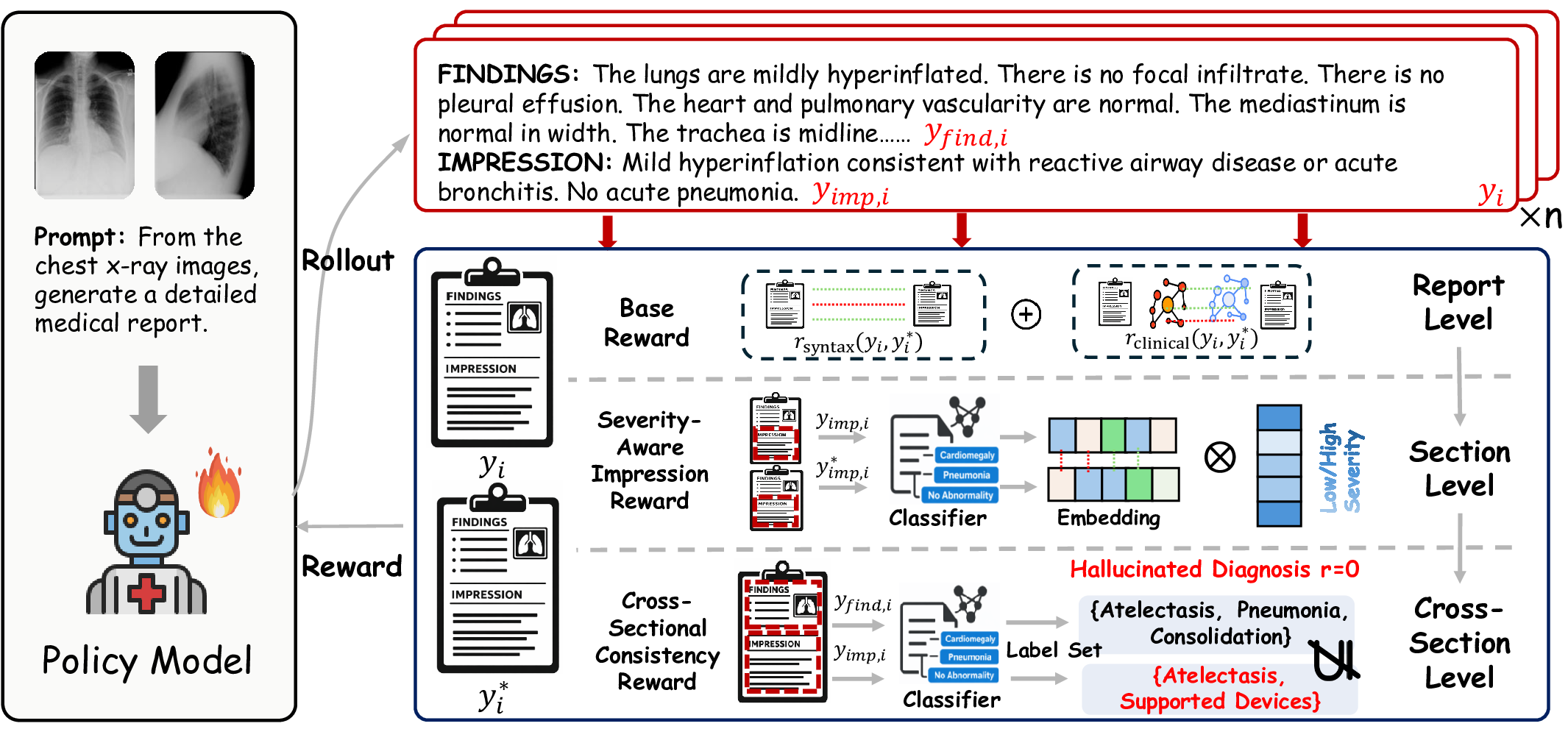} \caption{Overall architecture of \textbf{RadHiera}. The policy model generates $n$ rollouts from queries through GRPO optimization with the multi-level rewards: (1) base reward for policy optimization stability in whole report level, (2) Severity-Aware Impression Reward that leverages expert model embeddings to assess diagnostic severity levels, and (3) Cross-Sectional Consistency Reward that ensures alignment between Findings and Impression sections by detecting hallucinated diagnoses. This hierarchical reward enables generation of clinically effective and coherent radiology reports.}  \label{pipeline} \end{figure*}

\section{Method}

In this section, we first formulate the medical report generation task and introduce the reinforcement learning algorithm we build upon. Then we present the overall design of \textbf{RadHiera}, a semantic hierarchical reinforcement
learning framework as shown in Figure. \ref{pipeline}.

\subsection{Problem Formulation and Preliminaries}

Medical Report Generation (MRG) aims to generate structured and clinically coherent reports from medical images. 
A typical radiology report comprises two key sections: the \textit{Findings} section, which describes observable visual cues, and the \textit{Impression} section, which summarizes diagnostic conclusions.

Given an input medical image $x$ and its reference report $y^*=\{y_1^*,\dots,y_T^*\}$, a vision-language model $\pi_\theta$ models the conditional distribution of tokens:
\[
p(y_t \mid x, y_{<t}),
\]
where $y_t$ denotes the $t$-th generated token. Standard supervised fine-tuning (SFT) maximizes the likelihood of the reference report:
\[
\mathcal{L}_{\text{SFT}} = - \sum_{t=1}^{T}\log p(y_t^* \mid x, y_{<t}^*),
\]
While SFT ensures fluency and domain alignment, it fails to explicitly enforce clinical consistency between Findings and Impression or optimize clinical metrics, necessitating structured reinforcement learning.

To align model behavior with clinically grounded signals, we adopt Group Relative Policy Optimization (GRPO), a reinforcement learning algorithm that evaluates groups of candidate completions without a value network. For input $x$, the policy $\pi_{\text{old}}$ samples $G$ candidate reports $\{s_i\}_{i=1}^{G}$, each receiving a scalar reward $r_i$ from a reward function $R(s_i, x)$.

Within each group, rewards are normalized to compute relative advantage:
\begin{equation}
A_i=\frac{r_i-\mathrm{mean}(\{r_j\}_{j=1}^{G})}{\mathrm{std}(\{r_j\}_{j=1}^{G})+\epsilon},
\end{equation}
where $\epsilon$ is a constant. For each token $s_{i,t}$ in candidate $s_i$, the policy ratio is:
\begin{equation}
r_{i,t}(\theta)=\frac{\pi_\theta(s_{i,t}\mid x,s_{i,<t})}{\pi_{\text{old}}(s_{i,t}\mid x,s_{i,<t})}.
\end{equation}
The GRPO objective, incorporating clipping and KL regularization against a reference model $\pi_{\text{ref}}$, is:
\begin{equation}
\begin{split}
\mathcal{J}(\theta)=\mathbb{E}
\Bigg[
\frac{1}{G}\sum_{i=1}^{G}\frac{1}{|s_i|}\sum_{t=1}^{|s_i|}
\min\Big(
r_{i,t}(\theta)A_i,\,
\text{clip}(r_{i,t}(\theta),1-\epsilon,1+\epsilon)A_i
\Big)\\
-\beta D_{KL}\big(\pi_\theta(\cdot\mid x)\,\|\,\pi_{\text{ref}}(\cdot\mid x)\big)
\Bigg],
\end{split}
\end{equation}
where $\beta$ weights the KL penalty. This formulation increases the probability of above-average reports while preserving pretrained behavior. To optimizes policy behavior toward clinically structured reporting, we design a reward across different report semantic levels:

\begin{equation}
r(y,y^*) = r_{\text{base}}(y,y^*) + r_{\text{imp}}(y,y^*) + r_{\text{consist}}(y).
\end{equation}
Each reward targets a specific semantic granularity, detailed as follows.
\subsection{Report-Level Base Reward}

The base reward maintains general language quality and medical validity during policy optimization, while $r_{\text{hiera}}$ reshapes model behavior toward workflow-aligned reporting. The base reward is defined as
\begin{equation}
r_{\text{base}}(y,y^*) = r_{\text{syntax}}(y,y^*) + r_{\text{clinical}}(y,y^*).
\end{equation}
$r_{\text{syntax}}$ evaluates surface-level fluency and semantic similarity and $r_{\text{clinical}}$ measures factual correctness by extracting clinical entities from $y$ and comparing them with the  $y^*$. This base reward ensures that policy updates preserve medical validity and prevents significant deviations from the initial policy model, avoiding drift in language style and medical terminology. While the base reward maintains general quality, it does not explicitly encode the report structure. We therefore introduce other two semantic level rewards that shape the policy toward clinically meaningful structured reports.

\subsection{Severity-Aware Impression Reward}
Clinically, the Impression section reflects a high-level diagnostic decision. Errors on critical conditions (e.g., pneumothorax) carry higher clinical risk than minor lexical deviations. To model this asymmetry, we design a severity-aware reward on the Impression section.
Let $y^{\text{imp}}$ denote the generated Impression and $y^{\text{imp}*}$ the reference Impression. A frozen classifier \texttt{fcls} extracts binarized clinical labels:
\begin{equation}
z^{\text{imp}} = \texttt{fcls}(y^{\text{imp}}), 
\quad 
z^{\text{imp}*} = \texttt{fcls}(y^{\text{imp}*}).
\end{equation}
We compute a severity-weighted matching reward:
\begin{equation}
 r_{\text{imp}}(y,y^*) = \sum_{c} w_c \cdot \left( \mathbb{I}_{\text{TP}}^{(c)} - \mathbb{I}_{\text{FN}}^{(c)} - \mathbb{I}_{\text{FP}}^{(c)} \right),
\end{equation}
where $\mathbb{I}_{\text{TP}}^{(c)} = \mathbb{I}[z^{\text{imp}}_c = 1 \land z^{\text{imp}*}_c = 1]$ (and similarly for FN/FP), and $w_c$ assigns larger weights to clinically critical conditions \cite{cardinale2012critical_conditions_cxr,wang2020pulmonary_edema_severity}. This reward shapes the policy toward risk-sensitive diagnostic alignment and improves diagnostic accuracy.

\subsection{Cross-Sectional Consistency Reward}  
Radiology reports require \textit{Impression} conclusions to be grounded in \textit{Findings}. However, recent methods do not enforce this dependency, risking hallucinated or contradictory diagnoses. To address this, we extract clinical label sets from both sections:

\begin{equation}
z^{\text{find}} = \texttt{fcls}(y^{\text{find}}), 
\quad
z^{\text{imp}} = \texttt{fcls}(y^{\text{imp}}).
\end{equation}

Let $\mathcal{F}$ and $\mathcal{I}$ denote the sets of positive labels corresponding to $z^{\text{find}}$ and $z^{\text{imp}}$, respectively. And the consistency reward is defined as:
\begin{equation}
r_{\text{consist}}(y) 
= \begin{cases} 
1, & \text{if } \mathcal{I} \subseteq \mathcal{F} \\
0, & \text{otherwise}
\end{cases}
\end{equation}
Here, \( \mathcal{I} \subseteq \mathcal{F} \) checks if the \textit{Impression} set is a subset of the \textit{Findings} set, and the reward is 0 for any \textit{Impression} labels not found in the \textit{Findings} section (hallucinated diagnoses). This reward term shapes the policy to reduce cross-sectional contradictions and discourage unsupported diagnoses, ensuring that diagnostic conclusions in the \textit{Impression} are aligned with the \textit{Findings} section.

\section{Experiments}
\label{sec:experiments}

\noindent\textbf{Datasets} We evaluate our model on four datasets covering different imaging modalities:  CarotidUS-MRG: A private dataset of 12,000 carotid ultrasound studies with expert-annotated reports, split by patient (70/15/15\%) to prevent leakage. Reports follow a standardized format for structured clinical documentation.  MIMIC-CXR \cite{johnson2019mimic}: A public chest X-ray dataset with 377,110 frontal images and 227,827 associated reports. IU-Xray \cite{demner2016preparing}: A public chest X-ray dataset with 8,121 images and 7,470 reports from 3,955 patients. ReXGradient-160K \cite{zhang2025rexgradient}: The largest publicly available multi-site chest X-ray dataset, containing 273,004 images from 160,000 radiological studies across 109,487 patients from 79 medical sites. And we adopt the official split for consistency with prior studies \cite{zhang2024rexrankpublicleaderboardaipowered} enabling direct comparison with existing work.

\noindent\textbf{Evaluation Metrics.} We assess performance using standard natural language generation (NLG) and clinically grounded metrics. BLEU~\cite{papineni2002bleu}, BERTScore~\cite{zhang2019bertscore}, and SembScore~\cite{smit2020chexbert} evaluate linguistic quality. For MIMIC-CXR, IU-Xray, and ReXGradient, we follow the RexRank~\cite{zhang2024rexrankpublicleaderboardaipowered} protocol, incorporating RadGraph~\cite{jain2021radgraph}, $1/\text{RadCliQ-v1}$~\cite{yuEvaluatingProgressAutomatic2022}, CheXbert-F1\cite{smit2020chexbert}, and consistency (subset check). As for CarotidUS-MRG, where RadGraph and RadCliQ-v1 are inapplicable, we introduce KeyWordMatch, a metric based on a proprietary ultrasound labeler trained on expert-annotated data. It measures terminology fidelity by extracting and overlapping key medical terms. To ensure fair comparison, we fine-tune all baselines on IU-Xray, ReXGradient, and CarotidUS-MRG.

\noindent\textbf{Baselines and Implementation.} We evaluate our approach, which employs Supervised Fine-Tuning (SFT) on Qwen2.5-VL-3B \cite{bai2025qwen25vltechnicalreport} followed by GRPO with our semantic hierarchical rewards, against five state-of-the-art methods: R2GenGPT \cite{wang2023r2gengptradiologyreportgeneration}, MedGemma-4B~\cite{medgemma2024}, MedVersa\cite{zhou2025medversageneralistfoundationmodel}, CheXpertPlus~\cite{chambon2024chexpert}, and RadFM~\cite{wu2023generalistfoundationmodelradiology}. All models are trained under identical settings: learning rate $1\times10^{-5}$, LoRA rank 8, four generations per case and batch size 2. Training took approximately 672 GPU-hours on 8$\times$MetaX C500 GPUs (64,GB VRAM each).

\begin{table*}[!t]
\centering
\caption{
Performance comparison on MIMIC-CXR, IU-Xray and RexGradient. 
NLG metrics evaluate textual similarity at the report level. 
Clinical metrics assess factual and diagnostic correctness. 
 Bold = best, underline = second-best.
}

\small
\resizebox{\textwidth}{!}{%
\begin{tabular}{c|l|ccc|cccc}
\toprule
\multirow{2}{*}{Dataset} & \multirow{2}{*}{Model}
& \multicolumn{3}{c|}{NLG Metrics}
& \multicolumn{4}{c}{CE Metrics} \\

\cmidrule(lr){3-5} \cmidrule(lr){6-9}

& 
& BLEU-2 $\uparrow$ 
& BERTScore $\uparrow$ 
& SembScore $\uparrow$
& RadGraph $\uparrow$ 
& 1/RadCliQ-v1 $\uparrow$
& F1 $\uparrow$ 
& Consistency $\uparrow$ 
\\

\midrule
\multirow{9}{*}{MIMIC-CXR}

& RadFM~\cite{wu2023generalistfoundationmodelradiology}
& 0.081 & 0.281 & 0.245
& 0.111 & 0.625
& 0.245 & 0.741 \\

& CheXpert~\cite{chambon2024chexpert}
& 0.196 & 0.389 & 0.429
& 0.166 & 0.838
& 0.322 & 0.729\\

& R2GenGPT~\cite{wang2023r2gengptradiologyreportgeneration}
& 0.203 & 0.407 & 0.305
& 0.243 & 0.909
& 0.348 & 0.753\\

& MedGemma-4B~\cite{medgemma2024}
& 0.185 & 0.417 & 0.301
& 0.258 & 0.905
& 0.359 & 0.771\\

& MedVersa~\cite{zhou2025medversageneralistfoundationmodel}
& \underline{0.193} & \textbf{0.430} & \underline{0.315}
& \underline{0.273} & \underline{0.919}
& \underline{0.376} & \underline{0.776} \\

& \textbf{RadHiera (Ours)}
& \textbf{0.206} & \underline{0.421} & \textbf{0.442}
& \textbf{0.282} & \textbf{0.939}
& \textbf{0.394} & \textbf{0.851} \\

\midrule
\multirow{9}{*}{IU-Xray}

& RadFM~\cite{wu2023generalistfoundationmodelradiology}
& 0.228 & 0.491 & 0.589
& 0.254 & 1.320
& 0.421 & 0.841\\

& CheXpert~\cite{chambon2024chexpert}
& 0.264 & 0.516 & 0.608
& 0.257 & 1.619
& 0.469 & 0.828\\

& R2GenGPT~\cite{wang2023r2gengptradiologyreportgeneration}
& 0.271 & 0.558 & 0.601
& 0.269 & 1.751
& 0.483 & 0.852\\

& MedGemma-4B~\cite{medgemma2024}
& \underline{0.259} & 0.571 & 0.603
& 0.268 & 1.821
& \underline{0.502} & 0.891\\

& MedVersa~\cite{zhou2025medversageneralistfoundationmodel}
& 0.251 & \underline{0.579} & \underline{0.611}
& \underline{0.274} & \underline{1.852}
& 0.494 & \underline{0.898} \\

& \textbf{RadHiera (Ours)}
& \textbf{0.284} & \textbf{0.607} & \textbf{0.626}
& \textbf{0.313} & \textbf{2.325}
& \textbf{0.519} & \textbf{0.915}\\

\midrule
\multirow{9}{*}{ReXGradient}

& RadFM~\cite{wu2023generalistfoundationmodelradiology}
& 0.176 & 0.368 & 0.409
& 0.151 & 0.837
& 0.279 & 0.827\\

& CheXpert~\cite{chambon2024chexpert}
& 0.238 & 0.437 & 0.463
& 0.208 & 1.038
& 0.313 & 0.809 \\

& R2GenGPT~\cite{wang2023r2gengptradiologyreportgeneration}
& 0.251 & 0.478 & 0.479
& 0.231 & 1.050
& 0.334 & 0.812 \\

& MedGemma-4B~\cite{medgemma2024}
& 0.258 & \underline{0.486} & 0.482
& 0.242 & 1.120
& 0.341 &\underline{ 0.882} \\

& MedVersa~\cite{zhou2025medversageneralistfoundationmodel}
& \underline{0.266} & 0.481 & \underline{0.488}
& \underline{0.249} & \underline{1.170}
& \underline{0.349} & 0.874\\

& \textbf{RadHiera (Ours)}
& \textbf{0.298} & \textbf{0.544} & \textbf{0.492}
& \textbf{0.299} & \textbf{1.450}
& \textbf{0.376} & \textbf{0.891}\\

\bottomrule
\end{tabular}%
}

\label{tab:three_dataset_full}
\end{table*}

\begin{table}[t]
\centering
\caption{ Performance comparison the CarotidUS-MRG dataset. F1 is computed on the Impression section over 3 observation (stenosis, plaque and IMT thickness). RadGraph and RadCliQ-v1 are inapplicable, so we introduce KeyWordMatch, a metric based on a proprietary ultrasound labeler trained on expert-annotated data.}
\small
\resizebox{\textwidth}{!}{
\begin{tabular}{lcccccc}
\toprule
Model 
& BLEU-2 $\uparrow$
& BERTScore $\uparrow$
& SembScore $\uparrow$
& Keyword $\uparrow$ 
& F1 $\uparrow$ 
& Consistency $\uparrow$ 
\\
\midrule

R2GenGPT~\cite{wang2023r2gengptradiologyreportgeneration}
& 0.453
& 0.844
& 0.815
& 0.589 & 0.449 & 0.823\\

MedGemma-4B~\cite{medgemma2024}
& 0.458
& 0.846
& 0.812
& 0.571 & 0.461 & 0.873 \\

MedVersa~\cite{zhou2025medversageneralistfoundationmodel}
& \underline{0.473}
& \underline{0.868}
& \underline{0.837}
& \underline{0.593} 
& \underline{0.469} & \underline{0.885}
\\

\textbf{RadHiera (Ours)}
& \textbf{0.522}
& \textbf{0.915}
& \textbf{0.891}
& \textbf{0.742}
& \textbf{0.519} & \textbf{0.915}\\

\bottomrule
\end{tabular}
}
\label{tab:carotid_us_results}
\end{table}

\subsection{Main Results}

\noindent\textbf{Cross-Dataset Performance Superiority.}
RadHiera consistently outperforms others across all four datasets in Table \ref{tab:three_dataset_full} and \ref{tab:carotid_us_results}, leading in six of seven MIMIC-CXR metrics and sweeping IU-Xray, ReXGradient and the ultrasound CarotidUS-MRG. On MIMIC-CXR, while MedVersa slightly leads in BERTScore (0.430 vs. 0.421), RadHiera substantially improves semantic alignment with a SembScore of 0.442 vs. 0.315, alongside gains in clinical efficacy metrics like RadGraph (0.282 vs. 0.273) and Consistency (0.851 vs. 0.776). This gap widens on IU-Xray and ReXGradient, where RadHiera exceeds the runner-up by 0.473 in 1/RadCliQ-v1 (2.325 vs. 1.852), suggesting stronger performance to other datasets. On CarotidUS-MRG, consistent improvements across all metrics, including a F1 increase from 0.468 to 0.519, demonstrate robustness to the modality shift. 

\noindent\textbf{Diagnostic Consistency and Cross-Sectional Alignment.}
RadHiera excels at maintaining agreement between Findings and Impression sections, as reflected by the Consistency metric. Across MIMIC-CXR, IU-Xray, and ReXGradient, RadHiera scores 0.851, 0.915, and 0.891, surpassing second-best methods by 0.075, 0.017, and 0.009. These gains correlate with higher CheXbert-F1 on the Impression section (0.394 on MIMIC-CXR), linking enhanced cross-sectional coherence to more accurate diagnostic labeling. This consistent improvement extends to CarotidUS-MRG, as evidenced by the consistency (0.915 versus 0.885).

\begin{figure*}[t]
    \centering
    \includegraphics[width=\textwidth]{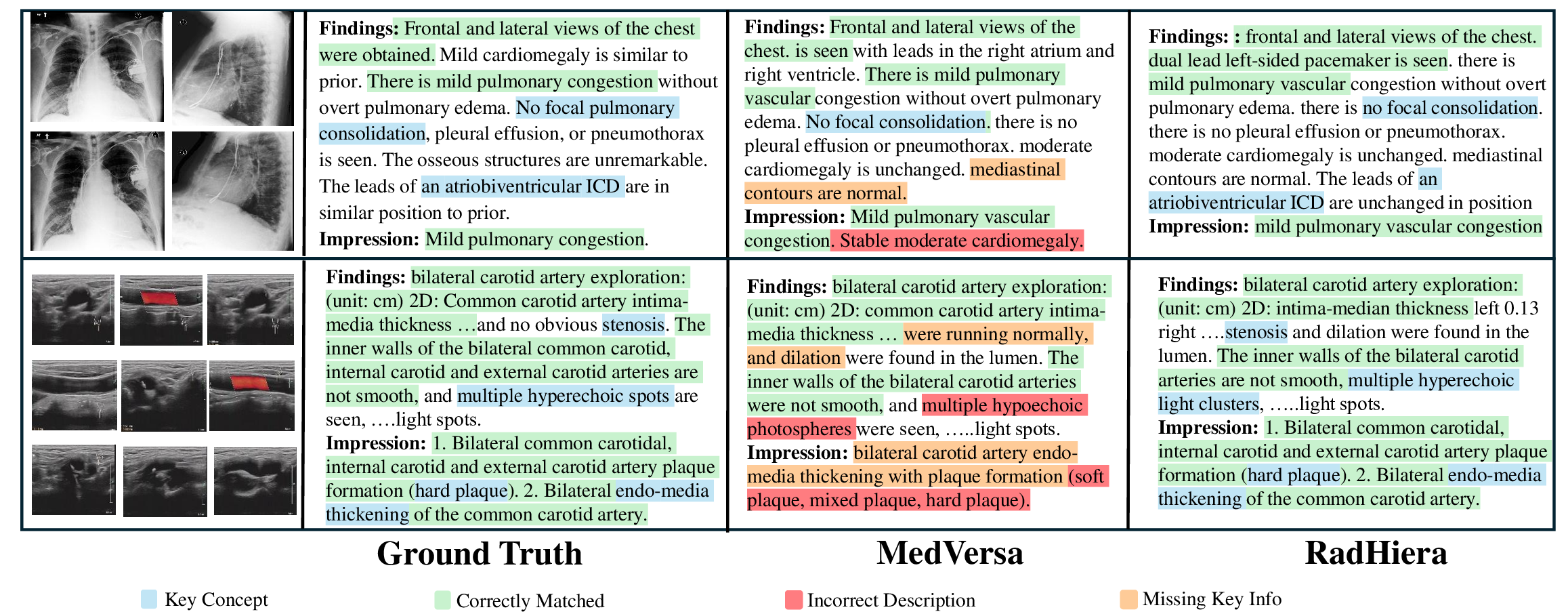}
    \caption{Comparison of MedVersa and RadHiera outputs against ground truth (GT). The figure highlights the differences between the reports from these models and GT. We observe that Medverse overlooked clinically critical indicator (ICD or teno-
sis severity ) and its Findings and Impression sections are sometimes weakly connected.In contrast, RadHiera generates clinically grounded reports with strong cross-sectional consistency and diagnostic accuracy.}
    \label{fig:qualitative}
\end{figure*}

\noindent\textbf{Case Study.}
Figure~\ref{fig:qualitative} compares MedVersa, RadHiera, and expert reports on MIMIC-CXR and CarotidUS-MRG. MedVersa often produces template-like descriptions focusing on superficial phrasing (e.g., pulmonary, intima-media thickness or plaque) while overlooking clinically critical indicators like ICD, stenosis severity or plaque characteristics. Its Findings and Impression sections are sometimes weakly connected, limiting diagnostic coherence. In contrast, RadHiera generates clinically grounded reports with strong cross-sectional consistency. It links imaging evidence to the diagnostic process, quantifies stenosis severity and characterizes plaque morphology. Resulting Impression sections are concise yet logically traceable to Findings, closely resembling expert-level summaries in structure and clinical	
reliability.

\subsection{Ablation Study}
\label{sec:ablation}

Experimental results in Table \ref{tab:ablation_RadHiera} demonstrate that RadHiera (Full) consistently outperforms the SFT baseline and standard GRPO across all dimensions. While standard GRPO improves SembScore from 0.402 to 0.411 and F1 from 0.332 to 0.368 over SFT, the full RadHiera configuration further elevates these to 0.442 and 0.394, respectively. Ablation reveals that consistency and Impression rewards serve complementary functions: removing the consistency reward reduces Consistency from 0.851 to 0.821 and RadGraph from 0.282 to 0.274, whereas ablating the Impression reward causes a sharper Consistency drop from 0.851 to 0.789. These findings indicate that while base GRPO provides a foundational gain, the joint optimization of hierarchical rewards yields synergistic improvements in both linguistic fluency and clinical structural alignment.
\begin{table*}[!t]
\centering
\caption{
Ablation study of reward components in RadHiera on MIMIC-CXR. 
We compare the full hierarchical reward against SFT baseline, standard GRPO with base reward, and partial reward combinations.
}
\label{tab:ablation_RadHiera}
\small
\resizebox{\textwidth}{!}{%
\begin{tabular}{l|ccc|cccc}
\toprule
\multirow{2}{*}{Model Configuration} 
& \multicolumn{3}{c|}{NLG Metrics} 
& \multicolumn{4}{c}{Clinical Efficacy (CE) Metrics} \\
\cmidrule(lr){2-4} \cmidrule(lr){5-8}
& BLEU-2 $\uparrow$ 
& BERTScore $\uparrow$ 
& SembScore $\uparrow$
& RadGraph $\uparrow$ 
& 1/RadCliQ-v1 $\uparrow$
& F1 $\uparrow$ 
& Cons. $\uparrow$ \\
\midrule

SFT (Qwen2.5-VL-3B) 
& 0.175 & 0.395 & 0.402 
& 0.241 & 0.901 & 0.332 & 0.747 \\ 

GRPO (Base Reward Only) 
& 0.188 & 0.402 & 0.411 
& 0.255 & 0.915 & 0.368 & 0.772 \\ 

RadHiera w/o Imp. Reward 
& 0.202 & 0.417 & 0.426
& 0.261 & 0.912 & 0.368 & 0.789 \\

RadHiera w/o Cons. Reward 
& 0.204 & 0.419 & 0.433
& 0.274 & 0.927 & 0.359 & 0.821 \\

\textbf{RadHiera (Full)} 
& \textbf{0.206} & \textbf{0.421} & \textbf{0.442}
& \textbf{0.282} & \textbf{0.939}
& \textbf{0.394} & \textbf{0.851} \\

\bottomrule
\end{tabular}%
}

\end{table*}
\section{Conclusion}

In this paper, we introduce RadHiera, a semantic hierarchical reinforcement learning framework that explicitly incorporates structural reporting constraints to improve the quality and consistency of medical report generation. By integrating a base reward, a severity-aware Impression reward and enforcing cross-sectional consistency between the Findings and Impression sections, RadHiera ensures clinically coherent and accurate diagnosis. Our extensive experiments on four datasets demonstrate that RadHiera significantly outperforms state-of-the-art methods, improving diagnostic accuracy and cross-sectional consistency.

\bibliographystyle{splncs04}
\bibliography{mybibliography}

\end{document}